# Kinematic analysis of a 3-UPU parallel Robot using the Ostrowski-Homotopy Continuation

Milad Shafiee-Ashtiani, Aghil Yousefi-Koma, Sahba Iravanimanesh, Amir Siavosh Bashardoust

Center of Advanced Systems and Technologies (CAST)
School of Mechanical Engineering, College of Engineering, University of Tehran, Tehran, Iran.
shafiee.a@ut.ac.ir, aykoma@ut.ac.ir, siravanimanesh@ut.ac.ir, as.bashardoust@ut.ac.ir

*Abstract*— The direct kinematics analysis is the foundation of implementation of real world application of parallel manipulators. For most parallel manipulators the direct kinematics is challenging. In this paper, for the first time a fast and efficient Homotopy Continuation Method, called the Ostrowski-Homotopy continuation method has been implemented to solve the direct and inverse kinematics problem of the parallel manipulators. This method has advantage over conventional numerical iteration methods, which is not rely on the initial values and is more efficient than other continuation method and it can find all solutions of equations without divergence just by changing auxiliary Homotopy function. Numerical example and simulation was done to solve the direct kinematic problem of the 3-UPU parallel manipulator that leads to 16 real solutions. Results obviously reveal the fastness and effectiveness of this method than the conventional Homotopy continuation methods such as Newton Homotopy. The results shows that the Ostrowski-Homotopy reduces computation time up to 80-97 % with more accuracy in solutions in comparison with the Newton Homotopy.

## I. INTRODUCTION

The parallel robotic manipulator has attracted the attention of many researchers and it also has growing applications in robotics, machine tools, positioning systems, measurement devices, and so on and even the other mechanism such as biped robots in double support phase behave like a parallel mechanism [5-6]. This popularity is a result of the fact that the parallel manipulators have more advantages in comparison to serial manipulators in many aspects, such as stiffness in mechanical structure, high position accuracy and high load carrying capacity [1]. However, they have some disadvantages such as limited workspace and complex forward position kinematics problems. In contrast to serial manipulators, the inverse position kinematics of parallel manipulators is relatively straightforward and the direct position kinematics is quite complicated [2]. It involves the solution of a system of coupled nonlinear algebraic equations that its variables describing platform posture and has many solutions [3].

Except in a limited number of parallel manipulators, there is no exact closed form solution [4]. So these nonlinear equations should be solved using numerical methods. Up to now, so many numerical methods has been developed to solve system of coupled nonlinear algebraic equations, such as the Newton–Raphson method which is efficient in the convergence speed [7]. Unfortunately, there always needs to guess the initial value in the iteration process. Good initial guess value may converge to solutions but bad initial guess value usually leads to divergence. Homotopy continuation method (HCM) is a type of perturbation and Homotopy method [8-9]. It can guarantee the answer by a certain path, if the auxiliary Homotopy function is chosen well. It does not have the drawbacks of conventional numerical algorithms, in particular the requirement of good initial guess values and the problem of convergence [10].

The recent development of the HCM was conducted by Morgan [11], Allgower [12]. Wu [8-10] introduced some techniques by combining Newton's and Homotopy methods to avoid divergence in solving system of coupled nonlinear algebraic equations. Also Wu [8], Varedi [13] and Abbasnezhad [14] have employed Newton-HCM in kinematics analysis of manipulators. Ostrowski's method was introduced by Alexander Markowich which has fourth order convergence and is an extension of Newton's method. [15, 16].

The latest advancement on Ostrowski's method was done by Grau [17], Chun [18] . Nor [19] presented some techniques by combining Ostrowski's and Homotopy Method to solve Polynomial Equations.

As it is clear, the numerical methods have the problem of expensive computational cost and for real-time application of robotics are not sufficient.

In this paper, for the first time the Ostrowski-HCM is employed to solve the direct and inverse kinematic of 3-UPU parallel manipulator with a very small computation time that allows for implementing kinematic analysis of parallel robots in real time applications. At the end the comparison was made between the Newton-HCM and Ostrowski-HCM results.

## II. THE HOMOTOPY CONTINUATION METHOD

Numerical iterative methods such as Newton-Raphson have two drawbacks. One is that bad initial guesses may leads to divergence and another is related to whether the iterative process will converge to desirable solutions. The HCM can rectify these deficiency [10]. In this algorithm we first write the Homotopy continuation equations with auxiliary Homotopy function and then solve this system of nonlinear equations by Newton-Raphson method and also with Ostrowski method.

Let us consider the following system of nonlinear equations:

$$F(x) = 0 \quad i.e. \begin{cases} f(x, y, \dots, z) = 0, \\ g(x, y, \dots, z) = 0, \\ \vdots \\ h(x, y, \dots, z) = 0, \end{cases} \quad (1)$$

Given a system of equation in n variables $x, y, \dots, z$ we modify the equations by omitting some of the terms and adding new ones until we have a new system of equations. The solution of this new system of equation may be easy to detect. We then modify the coefficients of the new system into the coefficient of the original system by a series of small increments and in last step we will solve the original system. This is called Homotopy continuation technique. In order to find the solution of Equations (1), we choose a auxiliary simple Homotopy function [8-10], as:

$$G(X) = 0, \quad (3)$$

The roots of G(X) must be easy to detect. Then, we can write the Homotopy continuation function as follows:

$$H(X, t) \equiv tF(X) + (1 - t)G(X) = 0, \quad (4)$$

$t$ is an arbitrary parameter and varies from 0 to 1, i.e., $t \in [0,1]$. Therefore, we have the following 2 boundary condition [8-10]:

$$H(X, 0) = G(X) \quad , \quad H(X, 1) = f(X) \quad (5)$$

Our goal is to solve the $H(X, t) = 0$ instead of $F(x) = 0$ by varying parameter $t$ by a series of small increments from 0 to 1. Hence numerical iteration formula of Homotopy Newton's method for solving these equations is as follows: as [10]:

$$\begin{bmatrix} \frac{\partial H_1(x_n, y_n, \dots)}{\partial x} & \frac{\partial H_1(x_n, y_n, \dots)}{\partial y} & \cdots \\ \frac{\partial H_2(x_n, y_n, \dots)}{\partial x} & \frac{\partial H_2(x_n, y_n, \dots)}{\partial y} & \cdots \\ \vdots & \vdots & \ddots \end{bmatrix} \begin{bmatrix} x_{n+1} - x_n \\ y_{n+1} - y_n \\ \vdots \end{bmatrix} = \begin{bmatrix} -H_1(x_n, y_n, \dots) \\ -H_2(x_n, y_n, \dots) \\ \vdots \end{bmatrix} \quad (6)$$

To rectify divergence problem, Wu [10] suggested some useful choice of auxiliary Homotopy function. They are polynomial, harmonic, exponential or any combinations of them.

*A. Ostrowski's Method*

Ostrowski's method was introduced by Alexander Markowich Ostrowski to find the roots of a single-variable nonlinear function. The method composed of two-step iterations using the following equations [15, 16]:

$$y_i = x_i - \frac{f(x_i)}{f'(x_i)} \quad (7)$$

$$x_{i+1} = y_i - \frac{f(x_i)}{f(x_i) - 2f(y_i)} \frac{f(y_i)}{f'(x_i)} \quad (8)$$

Where (7) is the conventional Newton's iterative formula while (8) is the Ostrowski's formula. Ostrowski's method which has fourth order convergence is an extension of Newton's method which has second order of convergence.

*B. Ostrowski Homotopy Continuation Method*

As we saw in previous section, there have been several studies that combine local methods with HCM such as Newton-HCM, Secant-HCM. Nor [19] combined local Ostrowski methods with HCM.
Ostrowski-HCM is as follows:

$$y_i = x_i - \frac{H(x_i)}{H(x_i)},$$
$$x_{i+1} = y_i - \frac{H(x_i)}{H(x_i) - 2H(y_i)} \frac{H(y_i)}{H(x_i)} \quad (9)$$
$$i = 0,1,2, \dots, k - 1$$

Where $i = 0,1,2, \dots, k - 1$ and t ∈ [0, 1]. To solve a system of system of coupled nonlinear algebraic equations, we have:

$$y_i = x_i - [D_x \boldsymbol{H}(x, t)]^{-1} \boldsymbol{H}(x, t)$$

$$x_i = y_i - [D_x \boldsymbol{H}(x, t)]^{-1} \frac{H(x_i) H(y_i)}{H(x_i) - 2H(y_i)} \quad (10)$$
$$i = 0,1,2, \dots, k - 1$$

Where $x_{i+1}, x_i$ and $\boldsymbol{H}(x_i, t)$ are vectors with dimensions n×1, $y_i = (y_1 \; y_2 \; \cdots \; y_n)^T$, $x_i = (x_1 \; x_2 \; \cdots \; x_n)^T$, and the $D_x \boldsymbol{H}(x, t) \; is$ Jacobian matrix of size n×n and the $H(x_i)$ and $H(y_i)$ are scalars.

### III. KINEMATICS MODEL OF AN OFFSET 3-UPU MANIPULATOR

Here we study kinematic analysis of offset 3-UPU (Universal–Prismatic– Universal) parallel manipulator with an equal offset in its six universal joints, as shown in Fig.1 [20]. 3-UPU manipulator has three prismatic limbs (legs) that link the base to the mobile platform by universal joints, and it can construct a 3-DOF pure translational motion [20]. Fig.3 shows a schema of the ith limb $i = (1,2,3)$, which links point B on the mobile platform and point A on the base by a passive universal (U) joint, an active prismatic (P) joint, and another passive universal (U) joint. So, it is called a 3-UPU parallel manipulator [20]. For the facility of kinematics analysis, three local frames, namely: $\sum_3: OX_3Y_3Z_3$, $\sum_2: OX_2Y_2Z_2 \; and \; \sum_1: OX_1Y_1Z_1$ are considered and attached to the base in order to describe the position and the first axis of a universal joint. Actually, frame $\sum_i (i = 1,2,3)$ is rotated along the X axis by angle $\beta_i$ from global frame $\Sigma$. $\beta_1 = 0$ since $\Sigma_1$ is chosen in a way to be parallel to $\Sigma$. Thus coordinates $\{X_i, Y_i, Z_i\}(i = 1,2,3)$ of a point in a local frames $\Sigma_i$ and its coordinates $\{X, Y, Z\}$ in the global frame have the following relation (Figs.2 and 4) [20]:

$$\begin{Bmatrix} X_i \\ Y_i \\ Z_i \end{Bmatrix} = \begin{bmatrix} 1 & 0 & 0 \\ 0 & \cos(\beta_i) & \sin(\beta_i) \\ 0 & -\sin(\beta_i) & \cos(\beta_i) \end{bmatrix} \begin{Bmatrix} X \\ Y \\ Z \end{Bmatrix} = \begin{Bmatrix} X \\ Y\cos(\beta_i) + Z\sin(\beta_i) \\ -Y\sin(\beta_i) + Z\cos(\beta_i) \end{Bmatrix} \quad (11)$$

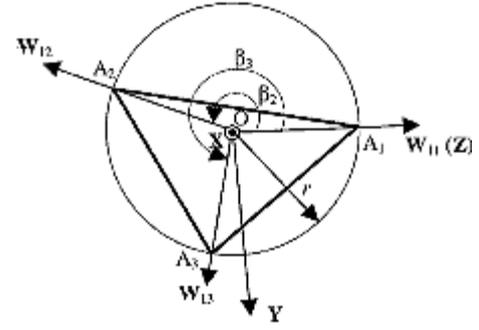

Figure 2. The base and global frame. [20]

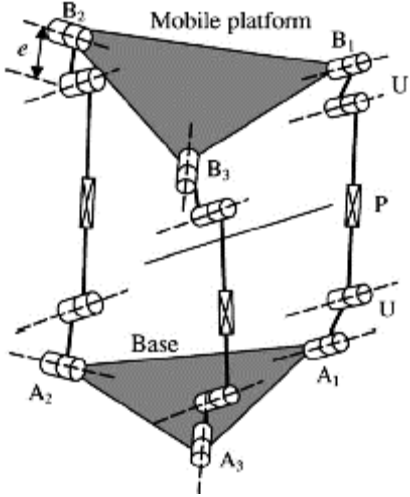

Figure 1. The 3-UPU parallel manipulator. [20]

*A. Inverse kinematics*

Let us consider the coordinates of $O_p$ be $(X, Y, Z)$ in the global frame $\Sigma$. For the inverse kinematics problem, the origin point $O_p(X, Y, Z)$ of the mobile platform in the global frame $\Sigma$ is known and the link lengths $L_i (i = 1,2,3)$ are need to be found. Since each branch limb can be regarded as a 3-DOF serial mechanism with its joint variables $\{\theta_{1i}, \theta_{2i}, L_i\}$. Point $O_p$ in a local frame $\Sigma_i (i = 1,2,3)$ can be represented as:

$$\begin{Bmatrix} X_i \\ Y_i \\ Z_i \end{Bmatrix} = \{A_i B_i\} + \begin{Bmatrix} 0 \\ 0 \\ r \end{Bmatrix} - \begin{Bmatrix} 0 \\ 0 \\ r_p \end{Bmatrix} \quad (12)$$

Where $\{X_i, Y_i, Z_i\}$ is the coordinates of $O_p$ in the local frame $\Sigma_i (i = 1,2,3)$ r is the circumradius of triangle $\Delta A_1 A_2 A_3$ in the base (Figs. 2 and 3) and $r_p$ is the circumradius of triangle $\Delta B_1 B_2 B_3$ with the circumcenter $O_p$ in the mobile platform (Figs. 3 and 4) [20]. From Fig.3, the vector from $A_i$ to $B_i$ in a local frame $\Sigma_i (i = 1,2,3)$ is:

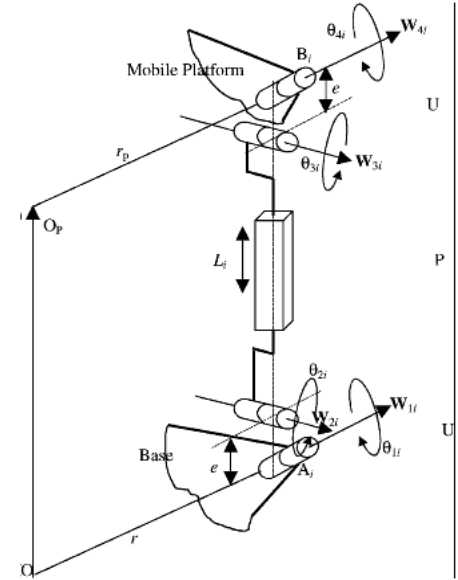

Figure 3. The ith limb of the manipulator [20]

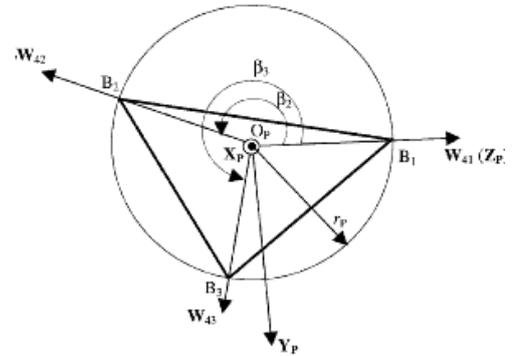

Figure 4. The mobile platform and its frame. [20]

$$\{A_i B_i\} = \begin{Bmatrix} (L_i \cos(\theta_{2i}) + 2e) \cos(\theta_{1i}) \\ (L_i \cos(\theta_{2i}) + 2e) \sin(\theta_{1i}) \\ L_i \sin(\theta_{2i}) \end{Bmatrix} \quad (13)$$

Where $L_i$ the length of ith prismatic joint, e is is the offset in a universal joint and $\theta_{1i}$ $\theta_{2i}$ are the rotation angles around its corresponding axis (Fig. 2) [20]. Substituting the value of $A_iB_i$ from Eq. (13) into Eq. (12), and some manipulation, yields [20]:

$$\begin{Bmatrix} X_i \\ Y_i \\ Z_i \end{Bmatrix} = \begin{Bmatrix} (L_i \cos(\theta_{2i}) + 2e) \cos(\theta_{1i}) \\ (L_i \cos(\theta_{2i}) + 2e) \sin(\theta_{1i}) \\ L_i \sin(\theta_{2i}) + dr \end{Bmatrix} \quad (i = 1,2,3) \quad (14)$$

Where $dr = r - r_p$

For finding the roots of this system of nonlinear equations we can use the Homotopy continuation method. Considering the structure parameters of the manipulator as: $dr = 0.382$ $e = 0.182$ $X_i = 0.5$, $Y_i = -1.5$ and $Z_i = 1$ we can rewrite equations (15) in the following simultaneous non-linear forms:

$$f = (L_i \cos(\theta_{2i}) + 2 \times 0.182) \cos(\theta_{1i}) - 0.5 = 0 \quad (15\text{-a})$$

$$g = (L_i \cos(\theta_{2i}) + 2 \times 0.182) \sin(\theta_{1i}) + 1.5 = 0 \quad (15\text{-b})$$

$$h = L_i \sin(\theta_{2i}) + 0.382 = 0 \quad (15\text{-c})$$

Therefore, we can write the Homotopy continuation function as follows:

$$H_1 = [(L_i \cos(\theta_{2i}) + 2 \times 0.182) \cos(\theta_{1i}) - 0.5] \times t + (1 - t) \times G_1 = 0 \quad (16\text{-a})$$

$$H_2 = [(L_i \cos(\theta_{2i}) + 2 \times 0.182) \sin(\theta_{1i}) + 1.5] \times t + (1 - t) \times G_2 = 0 \quad (16\text{-b})$$

$$H_3 = [L_i \sin(\theta_{2i}) + 0.382] \times t + (1 - t) \times G_3 = 0 \quad (16\text{-c})$$

We solve equations (16a)–(16c) by the Newton–Raphson and also by Ostrowski method and change the Homotopy parameter $t$ from 0 to 1. We choose the initial guesses of unknown parameters as: $(\theta_{1i0}, \theta_{2i0}, L_{i0}) = (-1, -0.5, 1.5)$

By appropriately regulating the auxiliary Homotopy function, we will obtain in total two sets of the roots of this problem without any divergence. The auxiliary Homotopy functions of these results are given in Table 1. The criterion used for measuring the iteration number of Ostrowski method in simulation is based on the both following terms simultanesly:

$$|(solution_{(newton-HCM)}) - (solution_{(ostrowski-HCM)})| < 10^{-6}$$
$$H_i(solution_{(newton-HCM)}) - H_i(solution_{(ostrowski-HCM)}) > 0$$

The results shows Ostrowski-HCM can reach the Newton-HCM accuracy with a quite less iteration and more accuracy. Based on results it's obvious that Ostrowski-HCM reduces Computation time for solving direct kinematic problem of the proposed manipulator up to 97 % in comparison with Newton-HCM.

*B. Forward Kinematics*

When three link lengths $L_i (i = 1,2,3)$ are known, the forward kinematics is to find $O_p(X, Y, Z)$. By eliminating $\theta_{1i}$ $\theta_{2i}$ Eq. (10) becomes an equation in terms of e, $dr$ and $L_i$:

$$[X_i^2 + Y_i^2 + (Z_i - dr)^2 - 4e^2 - L_i^2]^2 - 16e^2[(L_i^2 - (Z_i - dr)^2] = 0 \quad (17)$$

From equations (11) and (17), the following three equations can be exploited:

$$X^2 + Y^2 + (Z - dr)^2 - 4e^2 - L_1^2]^2 - 16e^2[(L_i^2 - (Z_i - dr)^2] = 0 \quad (18\text{-a})$$

$$X^2 + (Y \cos(\beta_2) + Z \sin(\beta_2))^2 + (-Y \sin(\beta_2) + Z \cos(\beta_2) - dr)^2 - 4e^2 - L_2^2]^2 - 16e^2[(L_2^2 - (-Y \sin(\beta_2) + Z \cos(\beta_2) - dr)^2] = 0 \quad (18\text{-b})$$

$$X^2 + (Y \cos(\beta_3) + Z \sin(\beta_3))^2 + (-Y \sin(\beta_3) + Z \cos(\beta_3) - dr)^2 - 4e^2 - L_3^2]^2 - 16e^2[(L_3^2 - (-Y \sin(\beta_3) + Z \cos(\beta_3) - dr)^2] = 0 \quad (18\text{-c})$$

Each of equations. (18a)–(18c) represents one equation in terms of X, Y, and Z. In order to solve the above equation, let us define:

$$P = X^2 + Y^2 + Z^2 \quad (19)$$

$$L_1 = 1.486, L_2 = 1.386, L_3 = 1.576, \beta_1 = 0, \beta_2 = \frac{2\pi}{3},$$
$$\beta_3 = \frac{4\pi}{3}, e = 0.182, dr = 0.382 \quad (20)$$

Substituting the value of P from Eq. (19) into equations (18a)–(18c), upon some simplifications and Substituting the values of the geometric parameters from (20), then we can write the Homotopy continuation function as follows

TABLE I. THE RESULTS OF HOMOTOPY CONTINUATION METHOD FOR INVERSE KINEMATICS

| No. | G1, G2, G3 | Method | $(\theta_1°), (\theta_2°), (L)$ | Iteration | Runtime (seconds) | Calculation Time Reduction% |
|---|---|---|---|---|---|---|
| 1 | $(\cos(\theta_{1i})), (\sin(\theta_{2i})), (L_i^2)$ | Ostrowski-HCM | (288.435), (26.91928), (1.36504) | 100 | 0.062951 | 97.735 |
| | | Newton-HCM | (288.434), (26.91909), (1.36504) | 100000 | 2.778823 | |
| 2 | $(-(\cos(\theta_{1i}))*2), (\sin(\theta_{2i})+2*\cos(\theta_{2i})+1), (L_i^2)$ | Ostrowski-HCM | (108.43492), (162.37417), (2.04095) | 290 | 0.069221 | 97.5467 |
| | | Newton-HCM | (108.43494), (162.37416), (2.04095) | 100000 | 2.821601 | |

$[(P - 0.764 \times Z - 2.194768)^2 - 1.170308549 + 0.529984 \times (Z - 0.382)^2] \times t + (1-t) \times G_1 = 0,$ (21-a)

$[(P + 0.382 \times Z + 0.661624 \times Y - 1.907568)^2 - 1.018097144 + 0.529984 \times (-0.866 \times Y - 0.5 \times Z - 0.382)^2] \times t + (1-t) \times G_2 = 0,$ (21-b)

$[(P + 0.382 \times Z + 0.661624 \times Y - 2.470348)^2 - 1.31636154 + 0.529984 \times (0.866 \times Y - 0.5 \times Z - 0.382)^2] \times t + (1-t) \times G_3 = 0$ (21-c)

The Homotopy parameter $t$ changes from 0 to 1 and the initial guesses of the unknown parameters are: $(P_0, Y_0, Z_0) = (1,1,1)$ We solve Equations (21a)–(21c) by the Ostrowski method and change the Homotopy functions $G_i (i = 1,2,3)$ to obtain the result. The auxiliary Homotopy functions and their results are given in Table 2. The simulation implemented in a hardware with the Intel® Core™ 2 Duo CPU and 4GB RAM in 64-bit Operating system by MATLAB program. we have found all of possible real solutions of direct kinematic of proposed parallel robot by this method. Ostrowski-HCM can reach to better accuracy with a quite less iteration than the Newton-HCM. Based on results it's clear that Ostrowski-HCM, with better accuracy, reduces computation time for solving direct kinematic problem of the proposed manipulator up to 80-97 % in comparison with Newton-HCM. Final 16 solutions by using Eq. (19) and result of Table.2 are listed in Table.3. The comparison of these results and the results reported by Wu [20] are in excellent agreement.

TABLE II.  THE RESULTS OF HOMOTOPY CONTINUATION METHOD FOR DIRECT KINEMATICS( INITIAL GIESS:$(P_0, Y_0, Z_0) = (1,1,1)$)

| Result No. | $G_1, G_2, G_3$ | Method | (P), (Y), (Z) | Iteration | Runtime (seconds) | Calculation Time Reduction% |
|---|---|---|---|---|---|---|
| 1 | P,Y,Z | Ostrowski-HCM | (1.57197), (0.71419), (0.58723) | 68 | 0.022847 | 92.2063 |
| 1 | P,Y,Z | Newton-HCM | (1.57197), (0.71419), (0.58723) | 10000 | 0.293147 | 92.2063 |
| 2 | (-P),Y,Z | Ostrowski-HCM | (2.37060), (0.93539), (-0.71988) | 143 | 0.0238 | 91.97367 |
| 2 | (-P),Y,Z | Newton-HCM | (2.37060), (0.93539), (-0.71988) | 10000 | 0.296524 | 91.97367 |
| 3 | P,(-Y),Z | Ostrowski-HCM | (1.17091), (-0.43286), (0.03721) | 253 | 0.031543 | 89.54744 |
| 3 | P,(-Y),Z | Newton-HCM | (1.17091), (-0.43286), (0.03721) | 10000 | 0.301773 | 89.54744 |
| 4 | (-P),(-Y),Z | Ostrowski-HCM | (1.54796), (-0.28820), (-1.08312) | 58 | 0.019755 | 93.18873 |
| 4 | (-P),(-Y),Z | Newton-HCM | (1.54796), (-0.28820), (-1.08312) | 10000 | 0.290034 | 93.18873 |
| 5 | (-P+5), (-Y-1), (-Z+3) | Ostrowski-HCM | (2.40438), (-0.28933), (1.34928) | 81 | 0.019237 | 93.4556 |
| 5 | (-P+5), (-Y-1), (-Z+3) | Newton-HCM | (2.40438), (-0.28933), (1.34928) | 10000 | 0.293946 | 93.4556 |
| 6 | (-P+5), (-Y-1), (Z+3) | Ostrowski-HCM | (3.21157), (-0.42875), (-0.02965) | 181 | 0.026975 | 90.71652 |
| 6 | (-P+5), (-Y-1), (Z+3) | Newton-HCM | (3.21157), (-0.42875), (-0.02965) | 10000 | 0.29057 | 90.71652 |
| 7 | (-P-1), (-Y+5), (Z+3) | Ostrowski-HCM | (2.48980), (-1.38318), (-0.64092) | 323 | 0.032144 | 88.89841 |
| 7 | (-P-1), (-Y+5), (Z+3) | Newton-HCM | (2.48980), (-1.38318), (-0.64092) | 10000 | 0.289544 | 88.89841 |
| 8 | (-P-1), (Y+5), (-Z-3) | Ostrowski-HCM | (1.47615), (-1.10076), (0.47274) | 791 | 0.078746 | 80.34387 |
| 8 | (-P-1), (Y+5), (-Z-3) | Newton-HCM | (1.47615), (-1.10076), (0.47274) | 10000 | 0.400618 | 80.34387 |

TABLE III. THE FINAL 16 SOLUTIONS OF THE DIRECT KINEMATICS PROBLEM

| Solution Number | P | X | Y | Z |
|---|---|---|---|---|
| 1 | 2.37060 | ±0.98863 | 0.93539 | -0.71988 |
| 2 | 1.47615 | ±0.20244 | -1.10076 | 0.47274 |
| 3 | 2.48980 | ±0.40720 | -1.38318 | -0.64092 |
| 4 | 3.21157 | ±1.73978 | -0.42875 | -0.02966 |
| 5 | 1.54719 | ±0.54013 | -0.28820 | -1.08312 |
| 6 | 1.57197 | ±0.84678 | 0.71419 | 0.58723 |
| 7 | 1.17091 | ±0.99103 | 0.43286 | 0.03721 |
| 8 | 2.40416 | ±0.70625 | -0.28929 | 1.34969 |

## IV. CONCLUSION

In this paper the Ostrowski-HCM is conducted to solve the forward kinematics of parallel manipulators. Analysis showed that that there are 16 real solutions for non-linear direct kinematic equations of 3-UPU manipulator. In result it has been proved that Ostrowski-HCM method reduces Computation time for solving direct kinematic problem and it is more effective and faster than conventional HCM methods. Also as shown in Table.4 solutions doesn't depend on initial guess. This method is so suitable for kinematic analysis of parallel robots and in particular in applications that fastness and accuracy of the method is important such as obstacle and singularity avoidance.

TABLE IV. SOME INITIAL GUESSES WHICH LEAD TO THE SAME ANSWER WITH COMMON HOMOTOPY FUNCTIONS IN TABLE 1 (NO.1)

| Number | $(\theta_{1,0}°, \theta_{2,0}°, L_0)$ | $(\theta_1°, \theta_2°, L)$ |
|---|---|---|
| 1 | (1,10,1.5) | (288.435), (26.91928), (1.36504) |
| 2 | (5729,85.94,0.5) | (288.435), (26.91928), (1.36504) |
| 4 | (-150, -81, 10) | (288.435), (26.91928), (1.36504) |
| 5 | (-0.05, 0.45, 100) | (288.435), (26.91928), (1.36504) |